\documentclass[letterpaper]{article} 
\usepackage{aaai25}  
\usepackage{times}  
\usepackage{helvet}  
\usepackage{courier}  
\usepackage[hyphens]{url}  
\usepackage{graphicx} 
\usepackage{natbib}  
\usepackage{caption} 
\usepackage{algorithm}
\usepackage{algorithmic}

\usepackage{newfloat}
\usepackage{listings}
\DeclareCaptionStyle{ruled}{labelfont=normalfont,labelsep=colon,strut=off} 
\floatstyle{ruled}
\newfloat{listing}{tb}{lst}{}
\floatname{listing}{Listing}
\title{Leveraging Group Classification with Descending Soft Labeling for \\Deep Imbalanced Regression}
\author{
    Ruizhi Pu\textsuperscript{\rm 1},
    Gezheng Xu\textsuperscript{\rm 1},
    Ruiyi Fang\textsuperscript{\rm 1},
    Bingkun Bao\textsuperscript{\rm 2},
    Charle X. Ling\textsuperscript{\rm 1}\thanks{Corresponding author}, 
    Boyu Wang\textsuperscript{\rm 1}\footnotemark[1]
}

\affiliations{
    \textsuperscript{\rm 1} Department of Computer Sceince, Western University\\
    \textsuperscript{\rm 2} School of Computer Science, Nanjing University of Posts and Telecommunications\\

}

\usepackage{bibentry}
\usepackage{amssymb,amsmath,amsthm}
\usepackage{mathtools}
\usepackage{amsfonts}
\usepackage{multirow}
\usepackage{diagbox}
\newtheorem{lemma}{Lemma}

\DeclarePairedDelimiter\floor{\lfloor}{\rfloor}

\begin{document}

\maketitle

\begin{abstract}
  Deep imbalanced regression (DIR), where the target values have a highly skewed distribution and are also continuous, is an intriguing yet under-explored problem in machine learning. 
  While recent works have already shown that incorporating various classification-based regularizers can produce enhanced outcomes, the role of classification remains elusive in DIR.
  Moreover, such regularizers (e.g., contrastive penalties) merely focus on learning discriminative features of data, which inevitably results in ignorance of either continuity or similarity across the data.
  To address these issues, we first bridge the connection between the objectives of DIR and classification from a Bayesian perspective. 
  Consequently, this motivates us to decompose the objective of DIR into a combination of classification and regression tasks, which naturally guides us toward a divide-and-conquer manner to solve the DIR problem.
  Specifically, by aggregating the data at nearby labels into the same groups, we introduce an ordinal group-aware contrastive learning loss along with a multi-experts regressor to tackle the different groups of data thereby maintaining the data continuity.
  Meanwhile, considering the similarity between the groups, we also propose a symmetric descending soft labeling strategy to exploit the intrinsic similarity across the data, which allows classification to facilitate regression more effectively.
  Extensive experiments on real-world datasets also validate the effectiveness of our method. 
\end{abstract}

%
\begin{links}
    \link{Code is available at}{https://github.com/RuizhiPu-CS/Group-DIR}
\end{links}

\section{Introduction}

Data imbalance exists ubiquitously in real-world scenarios, posing significant challenges to machine learning tasks as certain labels may be less observed than others or even missed during training.
Although the imbalanced problem has been extensively studied in the field of classification \cite{resam14},
how to tackle deep imbalanced regression (DIR) is still under-explored.

Due to the continuity of the label space and the dependence of data across nearby targets \cite{yang21devoling}, previous solutions in DIR primarily focused on estimating accurate imbalanced label density, such as label distribution smoothing (LDS), feature distribution smoothing (FDS) \cite{yang21devoling} and re-weighting \cite{reweight11, branco2017smogn}.
Meanwhile, \cite{ren2022balanced} proposed a balanced Mean Square Error (B-MSE) loss to accommodate the imbalanced distribution in the label space.
Recent works incorporated classification regularizers with the Mean Square Error (MSE) in DIR, such as contrastive regularization \cite{zha2023rank, keramati2023conr}, entropy regularization \cite{zhang2023improving}, and feature ranking regularization \cite{gong2022ranksim}, which have achieved significant performance improvements. 
Moreover, \cite{pintea2023step} formalized the relation between the balanced and imbalanced regression and empirically investigated how classification can help regression.

\begin{figure}
  \centering
  \includegraphics[width=0.75\columnwidth]{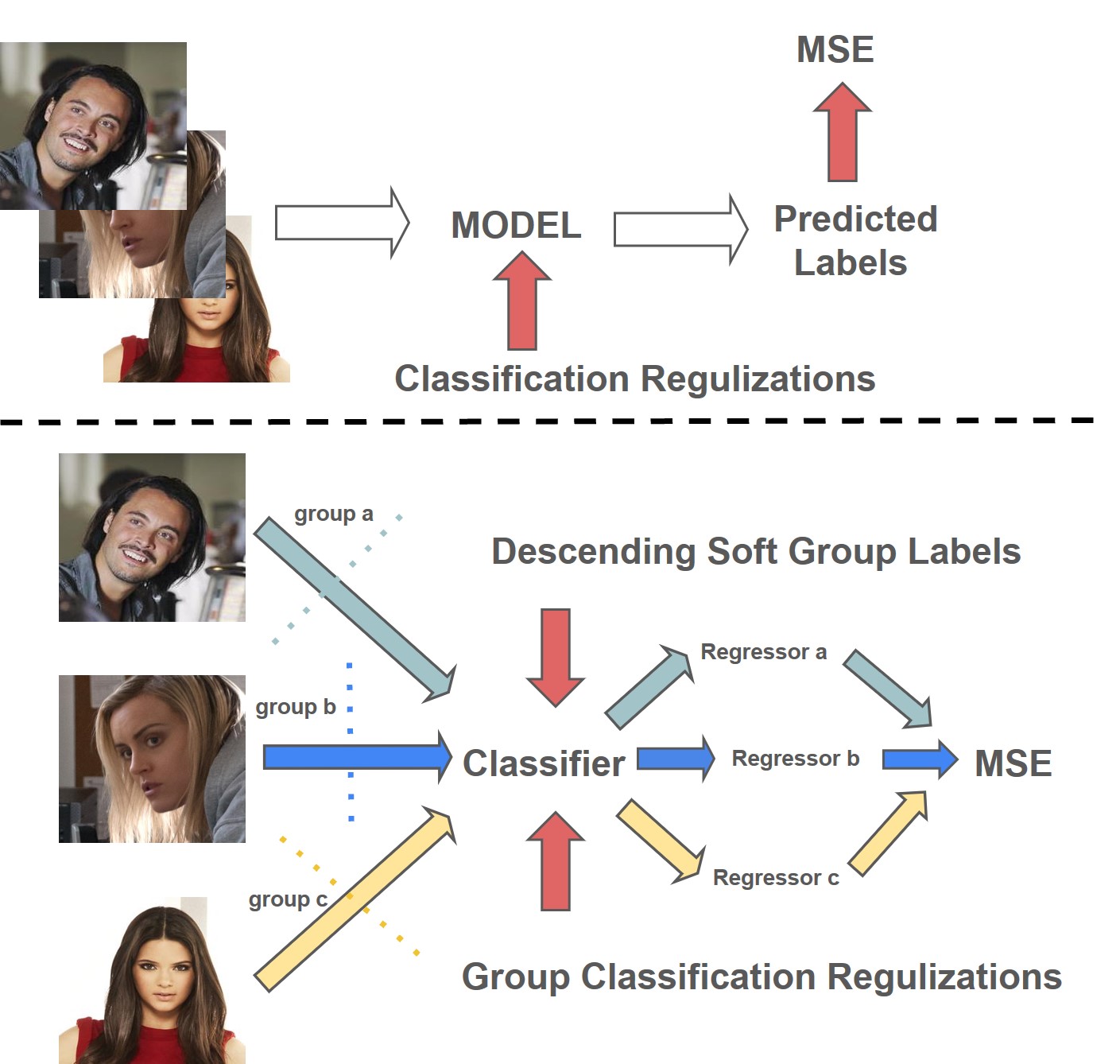}
  \caption{Comparison between previous works and ours. \textbf{Upper)} Previous methods directly incorporated classification regulations \cite{gong2022ranksim,zhang2023improving}.
  \textbf{Bottom)} We propose a descending soft labeling to leverage the classification to help DIR  (Different colors denote different groups of data).} \label{demo}
\end{figure}

Although incorporating classification regularizers in DIR has already achieved enhanced output,
the relationship between the objectives of classification and DIR remains elusive.
In the meantime, these classification regularizers would also force the model to focus more on the discriminative feature which is inappropriate for regression tasks.
For example, for a facial-image-based age regression task, the images corresponding to nearby labels exhibit both continuity and similarity. 
A photo of a 40-year-old person should resemble those of both 35-year-olds and 45-year-olds,  and also reflect an intermediate stage in age-related features. 
However, such property (data similarity) has been always ignored in existing classification-based methods.

In this paper, to investigate the connections between the classification and DIR, 
we revisit the objective of DIR from a Bayesian perspective.
We show that the objective of DIR can be decomposed into the combination of both group classification and sample regression within each group.
Inspired by this finding, we can explicitly leverage the classification to help DIR in a divide-and-conquer manner.

Specifically, considering that data with nearby labels would naturally be similar \cite{yang21devoling, pintea2023step} in DIR, we aggregate the data of close labels as the same groups.
Hereby, we divide the whole dataset into continuous but disjoint groups and convert the DIR into a classification problem.
In the meantime, these divided groups can not only preserve the ordinal information as their original labels but also provide us with a feasible way to explore the connection between the group classification and DIR.

Subsequently, since the decomposition of the DIR objective would also split the imbalance into both objectives of group classification and regression, 
inspired by \cite{liu2021self} that feature representations learned by self-supervised learning can exhibit imbalance-robust, we introduce an ordinal group contrastive learning to learn an ordinal high-quality feature representation to build a solid foundation for both classification and regression tasks.
Afterward, we make the group prediction for each learned representation on a classifier (Divide).
With this group estimation, we employ a multi-experts regressor to regress the representation on its corresponding predicted group (Conquer). 
The difference between our proposed method and the previous works can be found in Fig.\ref{demo}, where the previous works handle all data simultaneously while our work first divides the data into different groups and then conquers them with each expert regressor given their corresponding groups.

\begin{figure}[htbp]
    \centering
    \begin{minipage}[t]{0.48\textwidth}
    \includegraphics[width=8cm]{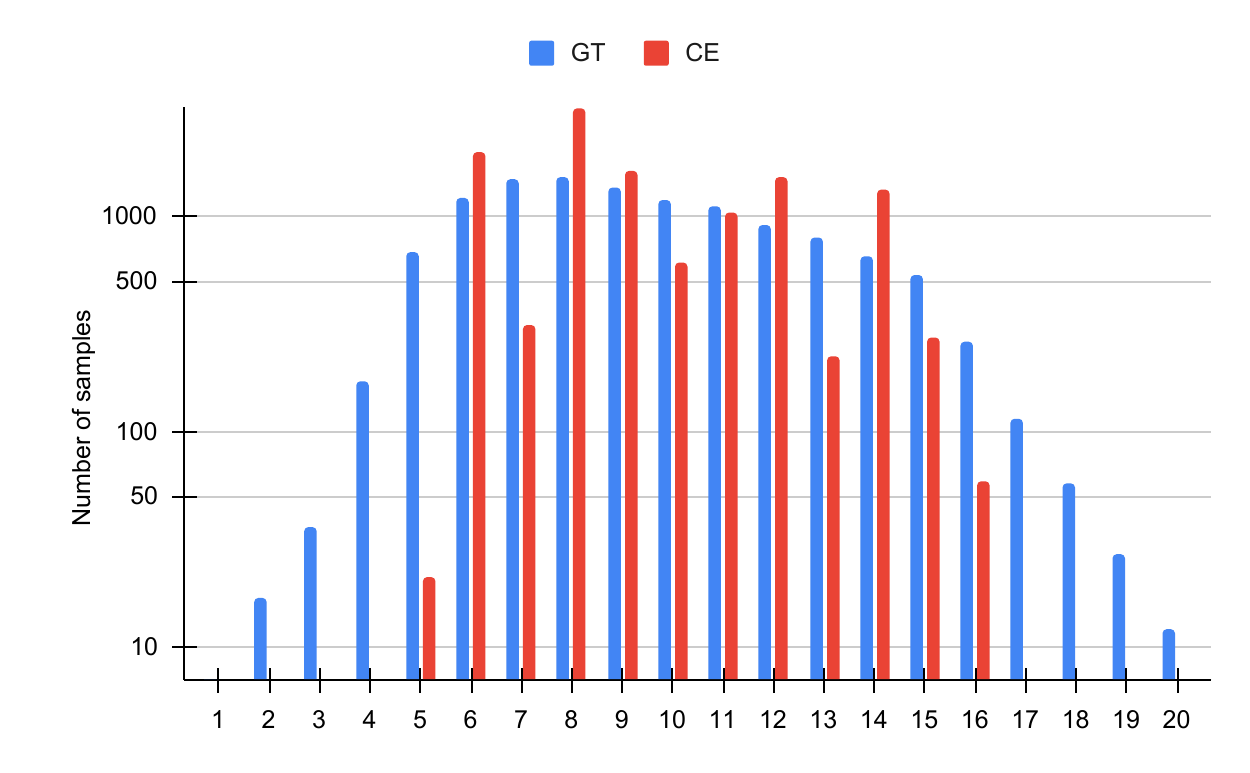}
    \caption{Comparison between the (Logarithm of) Ground Truth (GT) label and estimated label based on CE. X: groups.} \label{ob1}
    \end{minipage}
\end{figure}

However, empirical observation shows that it is difficult to make an accurate group estimation under standard classification loss such as cross-entropy (CE) loss.
For example, in Fig.\ref{ob1}, the data samples from group 1 to 5 (minorities) are rarely correctly predicted.
Instead, most of the data samples are over-estimated into group 6,8,12 and 14 (majorities).
The primary cause of this inaccurate prediction is the data dependence of nearby groups (images of close groups). \cite{yang21devoling} as each group also exhibits different levels of similarity between each other.
As shown in Fig.\ref{ob1}, groups with minority data samples would be easily misclassified into their neighboring groups with majority data samples.

As a result, these imprecise group predictions would misguide the data samples into the incorrect expert regressors and result in performance degradation.
To tackle this problem, we propose a symmetric descending soft labeling strategy that leverages the intrinsic label similarity of the data for the group prediction.
Since the labels can not only present the discrepancy information but also reflect the relative similarity between the data in DIR, we encode the group label into the soft labels which descend symmetrically from their group label until the end of the groups to capture the similarities between the groups.

In a nutshell, by incorporating the classification with the symmetric descending soft labeling into DIR, we provide a novel framework to address the DIR in a divide-and-conquer manner.
More importantly, we also conduct comprehensive experiments with various real-world datasets, demonstrating the effectiveness of our method.

In summary, we conclude our contributions as follows:
\begin{itemize}
    \item We revisit the objective of DIR from a Bayesian perspective, which motivates us to address the DIR in a divide-and-conquer manner.
    \item We incorporate an ordinal group-aware contrastive learning to learn a high-quality feature representation to provide a solid foundation for both classification and regression tasks in our decomposed objective.
    \item We introduce a multi-experts regressor to handle different groups of data with different expert regressors and we propose a symmetric descending soft labeling strategy to capture the similarity across the data in DIR.
\end{itemize}

\section{Motivation}
\subsection{Preliminary}
We study DIR in this paper.
In DIR, we assume that we have a training set $\{x_i, y_i\}_{i=1}^N$ with size $N$, where $x_i \in \mathbb{R}^d$ is the input with dimension $d$ and $y_i \in \mathbb{R}$ is the label.
Meanwhile, the distribution of this training set $p_{tr}$ is always highly skewed.
The objective of DIR is to learn a model from this highly skewed training set to generalize well on an unseen test set with the balanced distribution $p_{bal}$. 
In this paper, we aim to learn a feature extractor $f$ with parameter $\mathbf{w}_{f}$, a classifier $h$ with parameter $\mathbf{w}_{h}$ and a set of regressors $\varphi = [\varphi_{0},  \dots,\varphi_{|G|-1}]$ with parameter $\mathbf{w}_{\varphi} = [\mathbf{w}_{\varphi_0}, \dots,\mathbf{w}_{\varphi_{|G|-1}} ]$ simultaneously, the parameters of the model consists of $\theta=\{\mathbf{w}_{f},\mathbf{w}_{h},\mathbf{w}_{\varphi}\}$.

\subsection{Motivation}

We first revisit our goal from a Bayesian perspective. 
In DIR, our goal is to learn a model with parameter $\theta$ via a MSE loss to model the imbalanced training distribution $p_{tr}(y|x)$ and generalize well on the unseen balanced test distribution $p_{bal}(y|x)$.
Since directly adopting MSE loss in DIR is in fact to model the $p(y|x)$ for an underlying Gaussian distribution \cite{ren2022balanced}, a model learned from an imbalanced set would consequently underestimate rare labels,  limiting its ability to generalize to an unseen balanced set (Groups are mapped from labels, e.g., for a mapping $g=\floor*{\frac{y}{|G|}}$).

Therefore,  we can review the conditional distribution of training data $p_{tr}(y|x)$ as follows:
\begin{lemma}[Group-aware Bayesian Distribution Modeling for DIR] \label{motivation} The conditional distribution of  $p_{tr}(y|x)$ in the training of DIR can be decomposed into a combination of both classification and regression tasks summing over distinct groups: 
    \begin{align}
        \begin{split}
            p&_{tr}(y|x) = \frac{p_{tr}(x,y)}{p_{tr}(x)} 
            = \frac{\sum_{g\in G}p_{tr}(x,y, g)}{p_{tr}(x)} \\
            &= \frac{\sum_{g \in G}p_{tr}(g|x)p_{tr}(x)p_{tr}(y|x,g)}{p_{tr}(x)}  \\
            &= \sum_{g \in G}p_{tr}(g|x)p_{tr}(y|x,g) \nonumber
        \end{split}
    \end{align}
where $G$ is the set of groups, and $|G|$ is the number of groups, and we abbreviate $g$ as the group label.
\end{lemma}

\begin{figure}
  \centering
  \includegraphics[width=\columnwidth]{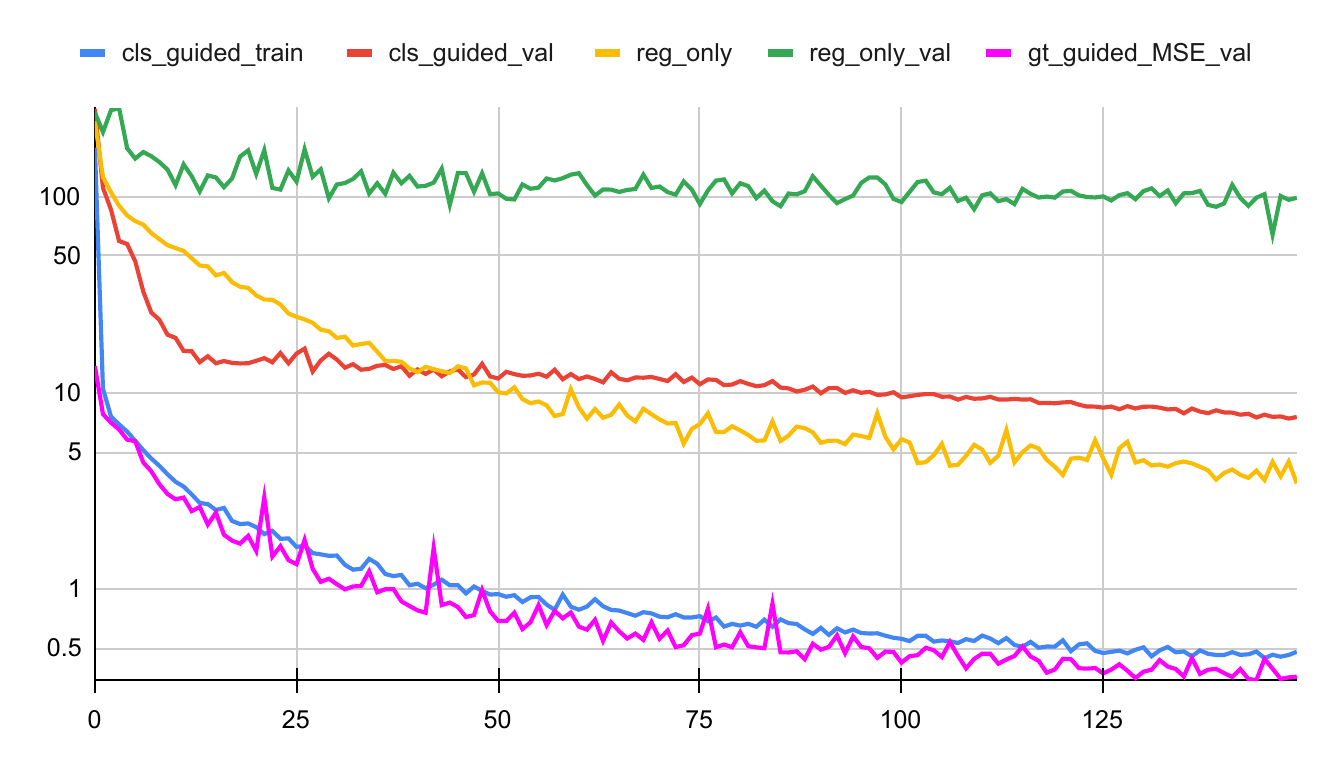}
  \caption{MSE results between model trained with MSE and model trained with the decomposition loss from Lem. \ref{motivation} (20 groups) on imbalanced Train \& balanced Validation set (AgeDB-DIR). Row: Epoch, Column: MSE (Note : column is in logarithmic scale for a more easier observation).} \label{mse_vs_ce}
\end{figure}

We take a step forward by taking negative logarithm at both sides \footnote{the inequality comes from $\log(a+b)\geq \log a + \log b$ for $0< a <1$ and $0< b < 1$ since the elements of logarithm in above are probabilities which less than 1, the inequality holds.}, we can obtain the learning objective of DIR in the form of loss as:
\begin{align}
    \begin{split}
        -\log &p_{tr}(y|x) \leq \sum_{g \in G} -\log(p_{tr}(g|x)p_{tr}(y|x,g)) \\
        &= \sum_{g \in G} \underbrace{-\log p_{tr}(g|x)}_{groups~classification} \underbrace{-\log p_{tr}(y|x,g)}_{labels~regression}
    \end{split}
\end{align}
\textbf{Remark:} The learning objective of DIR can be decomposed into two perspectives, 1) the objective of imbalance group classification to predict the group label, 2) the objective of imbalance regression to regress the data labels, showcasing that we can solve the DIR in a divide-and-conquer manner.
Empirical results from Fig.\ref{mse_vs_ce} also validates the effectiveness of objective decomposition in Lemma \ref{motivation} for addressing DIR.

As we can observe from Fig.\ref{mse_vs_ce}\footnote{The \_val denotes the validation performance, \_train denotes the training performance, cls\_guided denotes the data sample is regressed on the predicted regressor and gt\_guided denotes the data sample is regressed on the its true regressor.}, if we train a vanilla model with MSE loss only (regression-only), the training MSE loss curve and the validation MSE curve converges with different scales and the convergence speed diverges a lot, demonstrating that the training from the imbalanced set would result in unsatisfying results on the balanced validation set due to the imbalance in training set.

Instead, when we substitute the MSE loss with classification loss $p(g|x)$ (e.g. CE) and the classification-guided MSE loss $p(y|x,g)$ as Lemma \ref{motivation} , the classification-guided MSE loss exhibits a sharp converge compared with vanilla model (blue vs yellow). 
Moreover, the validation MSE of the classification-guided regression also converges more sharply compared with that of vanilla model (red vs green) and not even at the same scale, demonstrating that the representation learned from classification can help to address the DIR and showcasing the effectiveness of alleviating the negative impact of imbalance in DIR with our classification guided regression.
This motivates us to perform the divide-and-conquer by first estimating the groups $p(g|x)$ and then guiding the group-corresponding-regressors to regress the labels of the data samples $p(y|x,g)$.

Hereby, we connect the classification with the objective of the DIR (to model $p_{tr}(y|x)$ from $p_{tr}(y|x; \theta)$).
Furthermore, the above lemma also demonstrates that the objective of DIR can be upper-bounded by both classification and regression.
By minimizing the empirical risk of the classification of the groups (to model $p_{tr} (g|x)$ with $p_{tr}(g|x;\mathbf{w}_{f},\mathbf{w}_{h})$) and guiding the predictions of groups to minimize the empirical risk of the regression (to model $p_{tr}(y|x,g)$ with $p_{tr}(y|x, g;\mathbf{w}_{f},\mathbf{w}_{\varphi})$) simultaneously, we can properly address the DIR from a Bayesian perspective.
More importantly, this motivates us to solve the DIR problem in a divide-and-conquer manner as we can leverage the group classification $p(g|x)$ to guide the learning of regression $p(y|x,g)$.

\section{Methodology}
In this section, we introduce an ordinal group-aware constrastive learning to learn a high-quality feature representation which is beneficial for both classificaiton and regression. 
Then, we leverage a multi-experts regressor to conduct regression under the guidance of the group predictions to fully exploit the benefits from classification to help regression in a divide-and-conquer manner. 
Furthermore, we propose a symmetric descending soft labeling strategy to capture the data similarity across groups.

\subsection{Ordinal Group-aware Contrastive Learning}
As in DIR, label space is not only continuous but also ordinal.
Consequently, we introduce an ordinal group-aware contrastive learning to learn a high-quality feature representation. 
Meanwhile, this high-quality representation can also act as a solid foundation for both classification $p(g|x)$ and the regression $p(y|x,g)$ tasks as described in our objective decomposition from Lemma \ref{motivation}.

Inspired by \cite{zha2022supervised,XIAO2023119669}, we introduce an ordinal contrastive learning in a group-aware manner.
Since data samples at nearly labels would have similar features (e.g. facial samples from 30 to 40 would be similar with each other), we cluster the data samples with their corresponding groups.
Different from \cite{zha2022supervised}, we concentrate on investigating relationships between these groups to help to learn a high-quality feature representation.

As these distinct groups would preserve the ordinal as their original labels (e.g. the label of arbitrary sample in group 0 would always be smaller than the label of arbitrary sample in group 1), we focus on constructing an ordinal group-aware contrastive learning framework where the learned feature representations can also preserve this ordinal characteristics between the groups.

In order to achieve this goal, for an anchor group label $i$ and another arbitrary group label $j$, we push away other samples whose group label distance are more distant than $i$ and $j$.
If two samples are in the same group, then we pull them together at the feature space. 
In this way, data samples with different distances in group labels would be pushed away in different degrees, as the close groups would be pushed less than the distant groups.

Hereby, we formulate the ordinal group-aware contrastive loss as the following:
\begin{align}
    \begin{split}\small
        \mathcal{L}_{grc}(\mathbf{w}_{f}) = -\frac{1}{B(B-1)}\sum_{i=1}^{B} \sum_{\substack{j=1,\\j\neq i}}^{B}  \log\frac{s(z_i,z_j)}{\sum_{k=1}^{B}\mathbf{1}_{[\phi(i,j,k)] } s(z_i, z_k)}  \label{con} 
    \end{split}
\end{align}
where for the index $i,j,k$ of three arbitrary data samples in a batch, $z$ is the feature representation, $s(i,j)$ is the abbreviate of $exp(sim(z_i, z_j)/t)$ and $sim(\cdot)$ denotes the similarity function (e.g. cosine similarity), $exp(\cdot)$ is the exponential function,
$\phi(i,j,k) \triangleq \left\{ k\neq i, d(g_i, g_k) \geq d(g_i, g_j) \right\} $ is the condition of the zero-one indicator $\mathbf{1}$ (return 1 where $\phi$ satisfies and 0 vice verse), $g$ denotes the group label of the data sample, $t$ is the temperature hyper-parameter, $B$ is the batch size, and $d(\cdot)$ denotes the distance function (e.g. L1 distance).
By comparing the relative distance of group labels between arbitrary two samples, we can achieve the group ordinal as that of the labels in the feature space.

\subsection{Classification guided multi-experts regression : modeling $p(y|x,g)$}
With the acquired contrastive representations, we introduce a multi-experts regressor to tackle each group of data in a divide-and-conquer manner.
At the training phase, given the ground truth group label of each data sample, we conduct regression on its corresponding expert regressor.
At the testing phase, each data sample is first classified into a group.
Since each predicted group is in correspondence with an expert regressor.
Then, we conduct regression on the predicted expert regressor.

Therefore, we formulate the multi-expert regression MSE loss as follows:
\begin{equation}
    \mathcal{L}_{mse}(\mathbf{w}_{f}, \mathbf{w}_{\varphi}) = \sum^{|G|-1}_{g=0, y\in [g]} (y_{\varphi_g} - \hat{y}_{\varphi_g} )^2 
\end{equation}
where $y\in [g]$ denotes the label $y$ belongs to group $g$.
We have $\hat{y}_{\varphi_g} = \mathbf{w}_{\varphi_g}(z_{\varphi_g})$ for the group of data samples whose ground truth group labels are $g$ and their learned representations $z$ are then forwarding to their corresponding regressors $\varphi_g$ to obtain the prediction $\hat{y}_{\varphi_g}$.
We abbreviate the ground truth target labels as $y_{\varphi_g}$.
Moreover, LDS can also be utilized to further tackle the intra-group imbalance.
Since the final MSE is calculated on each data sample and each data sample corresponds to each group, 
we accumulate the MSE loss over all groups.

\subsection{Symmetric Descending Soft Labeling for group classification : modeling $p(g|x)$} 

However, since the nearby label data would exhibit data dependence \cite{yang21devoling} in DIR, in our framework, the nearby group data would also exhibit data dependence.
Consequently, the data dependence of nearby groups and inherent group imbalance would hinder us from making accurate group estimations for regression.

\begin{figure}[htbp]
    \begin{minipage}[t]{0.48\textwidth}
    \centering
    \includegraphics[width=8cm]{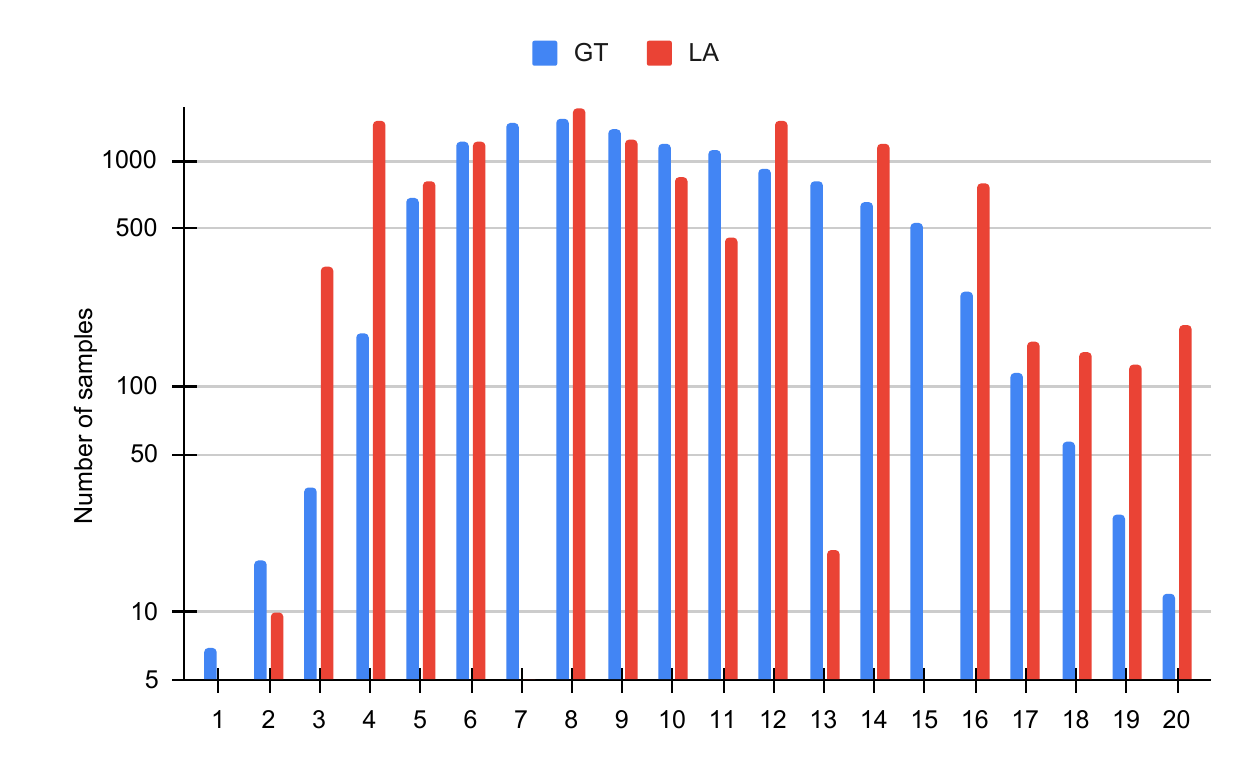} 
    \caption{Comparison between the (Logarithm of) Ground Truth (GT) label and estimated label based on LA. X: groups.} \label{ob2}
    \end{minipage}
\end{figure}

When we directly utilize the standard CE loss to estimate the group label of the data, as can be observed from Fig.\ref{ob1}, the predictions mostly fall into the groups with the majority of samples.
Meanwhile, when we adopt logits adjustment (LA) \cite{menon2021longtail}, which is one of the most effective imbalance classification solutions, to predict the group labels, another empirical observation arises in Fig.\ref{ob2} that this method over-estimate the groups with minority samples.

Therefore, empirical results in Fig.\ref{ob1} and Fig.\ref{ob2} have shown that classification loss such as CE and LA perform poorly in group prediction of DIR.
More importantly, as can be observed from Fig.\ref{absolute}, in both CE and LA, the data dependence of the groups would lead the predictions to mainly fall into nearby groups (high absolute difference of misclassification in Fig.\ref{absolute}).
The reason for this is the classification solutions would focus on the discriminative information (as we stated above) while ignoring the data similarities across the groups.

In standard classification loss such as  CE, the ground truth for one group label $g$ is encoded as a vectorized label $l_{gt} = [\dots,0, 0, 1, 0, 0, \dots]$, where 1 is at the position of $g$-th index, and 0 is at the rest indexes and the CE loss $\mathcal{L}_{ce} = -\log p^g$ is only calculated on the prediction at the index $g$ for the group prediction $p = [p^0, \dots, p^g, \dots, p^{|G|-1}]$.
As a result, only the information on the ground truth index is provided (set to 0) while others are all overlooked (set to 0).
However, since the group label is not merely continuous, it's also essential to recognize that labels within DIR encode intrinsic relative similarities between them.
Thus, directly adopting classification loss in group prediction is not applicable in DIR.

Therefore, we introduce a symmetric descending Soft labeling strategy into the group classification to fully exploit the similarity nature of the groups.
To convert a scalar group label into a vectorized label for training, for a group with ground truth label $g$, we assign the $g$-th index in the label vector with the highest value of $|G|$ and decrease it symmetrically from the position of the current index until the end.
Thus, the soft label of the scalar group label $g$ would be encoded as  $l_{soft} = [\dots, |G|-2\beta, |G|-\beta, |G|, |G|-\beta, |G|-2\beta, \dots]$, where, $|G|$ is at the index of $g$ in the label vector, $\beta$ is a hyper-parameter e.g. $\beta=1$ and it denotes the relative distance between two neighboring labels.
We formulate the soft label $q_{soft}$ of a data sample from the ground truth group label as:
$q_{soft} = \sigma(l_{soft})$ 
where $\sigma$ denotes the SoftMax function, as $\sigma(q_i) = \frac{e^{q_i}}{\sum_{j=1}^{|G|}e^{q_j}}$.
Moreover, we briefly show two extreme cases for our soft label,  in the case when $g=0$, the $l_{gt\_soft}= [|G|, |G|-1, |G|-2, \dots,1]$, and in the case when $g=|G|$, the $l_{gt\_soft}=[1,\dots, |G|-2, |G|-1, |G|]$.

The soft label cross-entropy loss for a data sample with group label $g$ (corresponding with the regressor $g$) in a batch $B$ as the following:
\begin{equation}
    \mathcal{L}_{soft}(\mathbf{w}_{f}, \mathbf{w}_{\varphi_g}) = \sum_{j=1}^B\sum_{g=0}^{|G|-1}  q_j^g\log p_j^g
\end{equation}
where $p_j^g$ denotes group prediction and $q_j^g$ is the soft label in $q_{soft}$ of sample $j$ at index $g$.

By encoding the ground truth labels into soft labels, we can preserve the relative group information of all groups in one single label, providing comprehensive data information for the group classification and also contributing to the regression.
T-SNE of the different classification criteria (Soft Labeling/CE/LA) shows the effectiveness of our method. %

\subsection{Final Loss}
By aggregating the above losses together, the final objective of our proposed method is :
\begin{align}
    \mathcal{L}_{final} = \mathcal{L}_{grc}  +  \lambda_1\mathcal{L}_{mse} + \lambda_2\mathcal{L}_{soft}  
\end{align}
where $\lambda_1$ and $\lambda_2$ are hyper-parameters to balance the losses.
 
\section{Experiments}\label{sec:exp}
\subsection{Datasets}
We validate our proposed method with the following real-world dataset which includes both visual tasks and natural language processing task:

\textbf{IMDB-WIKI-DIR} is a large-scale real-world human facial dataset constructed by \cite{imdb_wiki} and re-organized for imbalance tasks by \cite{yang21devoling}, it contains 235K face images.
There are  191.5K imbalance training images, 11K balanced validation images, and 11K balanced test images. 
The dataset was manually divided given the bin length of 1 year (each bin can be regarded as the target label as in \cite{yang21devoling}).

\textbf{AgeDB-DIR} is another real-world human facial dataset constructed by \cite{agedb} and also re-organized by \cite{yang21devoling}. 
It contains 12.2K images training data, 2.1K images validation data, and 2.1K images test data.
The bin length is also 1 year but the minimum age is 0 and the maximum age is 101.

\textbf{STS-B-DIR} is a text similarity score dataset constructed by \cite{sts} and re-constructed by \cite{yang21devoling}. 
It is collected from news headlines, videos, image captions, and natural language inference data. 
The dataset is a set of sentence pairs annotated with an average similarity score, and the range of scores varies from 0 to 5. 
There are 5.2K pairs for the training, 1K balanced pairs for validation, and 1K balanced pairs for test.
Each bin length is 0.1.

\subsection{Implementation Details} 
\textbf{Baselines and experiment set up} 
We conducted our experiments with the backbone based on ResNet-50 for AgeDB-DIR \& IMDB-WIKI-DIR dataset.
For STS-B-DIR, we follow the same standard experiment setting as in \cite{yang21devoling, ren2022balanced}, we adopted the BiLSTM + GloVe word embeddings and preprocessed them in the experiment.
Moreover, we follow the training procedures and hyper-parameters (e.g., temperature $t$) as \cite{zha2023rank}, but apart from \cite{zha2023rank} which only used a sub-sample of both datasets (e.g., 32K for IMDB-WIKI-DIR), we stick to the setting of \cite{yang21devoling} and use the full training set with the batch size of 128 for training.
Same as \cite{yang21devoling, branco2017smogn}, the train data distribution is always highly skewed while the test distribution is balanced.
More details can be found in the Appendix.

\renewcommand{\floatpagefraction}{.9}
\begin{figure}
  \centering
  \includegraphics[width=0.47\textwidth]{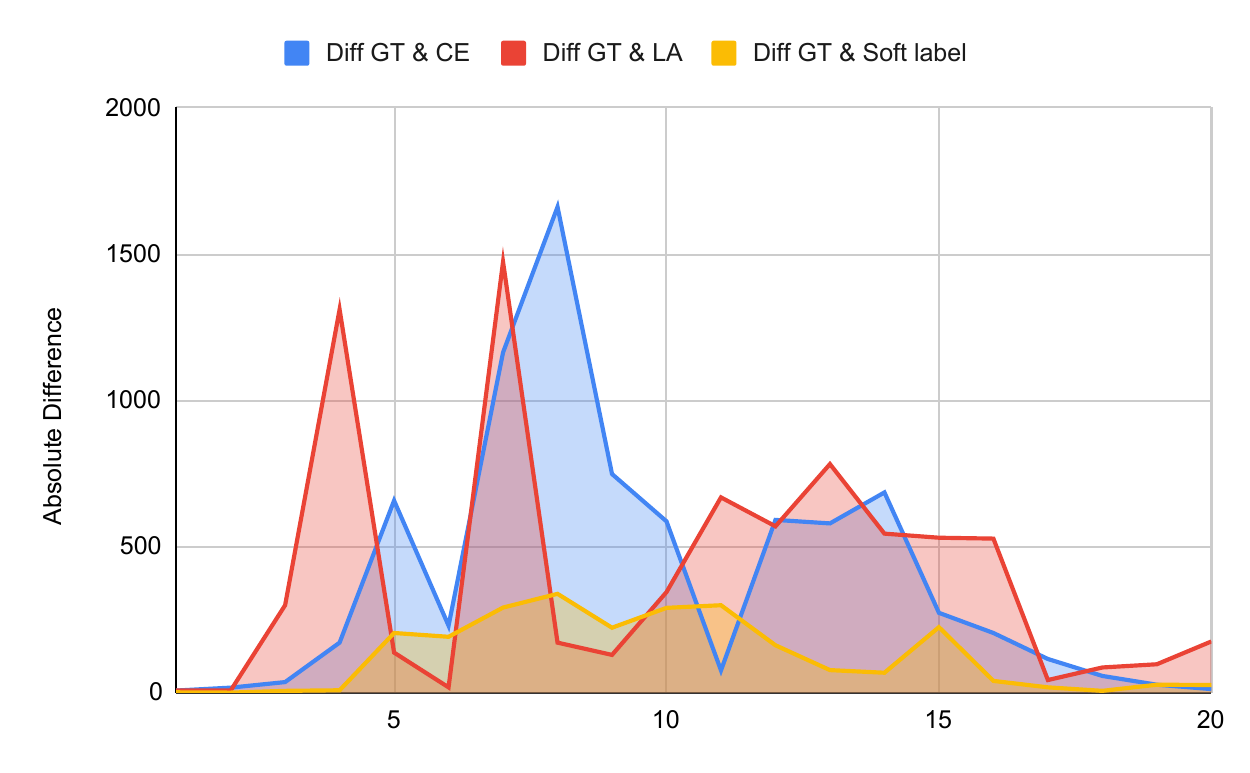}
  \caption{Comparison of the absolute difference (Diff) between group predictions and ground truth in three methods (CE, LA, and Ours on AgeDB-DIR). Lower denotes the more accurate group predictions. X-axis : group numbers.} 
  \label{absolute}
\end{figure}

\subsection{Result Analysis}
\textbf{AgeDB-DIR} : 
In the dataset AgeDB-DIR, it is obvious that our method outperforms other methods in Tab.\ref{agedb_tabel}.
In particular, we show that our method can better deal with the majority and median, without greatly sacrificing the performance of the minorities as in the previous works.
Compared with existing works, our work achieved a state-of-art (SOTA) performance with an MAE of 6.87.
Meanwhile, our work has lower GM,  which shows that our results of MAE are averagely smaller than other works, showing the effectiveness of our proposed method.

We show that in Fig.\ref{absolute},
we use the absolute difference between the ground truth labels and the estimated labels to identify if our proposed method can help the classification (how accurate the group estimation can be given the ground truth).
Our proposed symmetric descending soft labeling significantly outperforms others in group estimation (compared with Fig.\ref{ob1} \& \ref{ob2}), that is because the soft labels can help the representations to fully exploit the similarity characteristics of the data from the descending labels. 
Consequently, it contributes to a more accurate group estimation than other existing works, resulting in minimizing the $\log p(g|x)$ and the gap $\Delta$ at the same time.

\begin{figure}[htbp]
    \centering
    \includegraphics[width=0.48\textwidth]{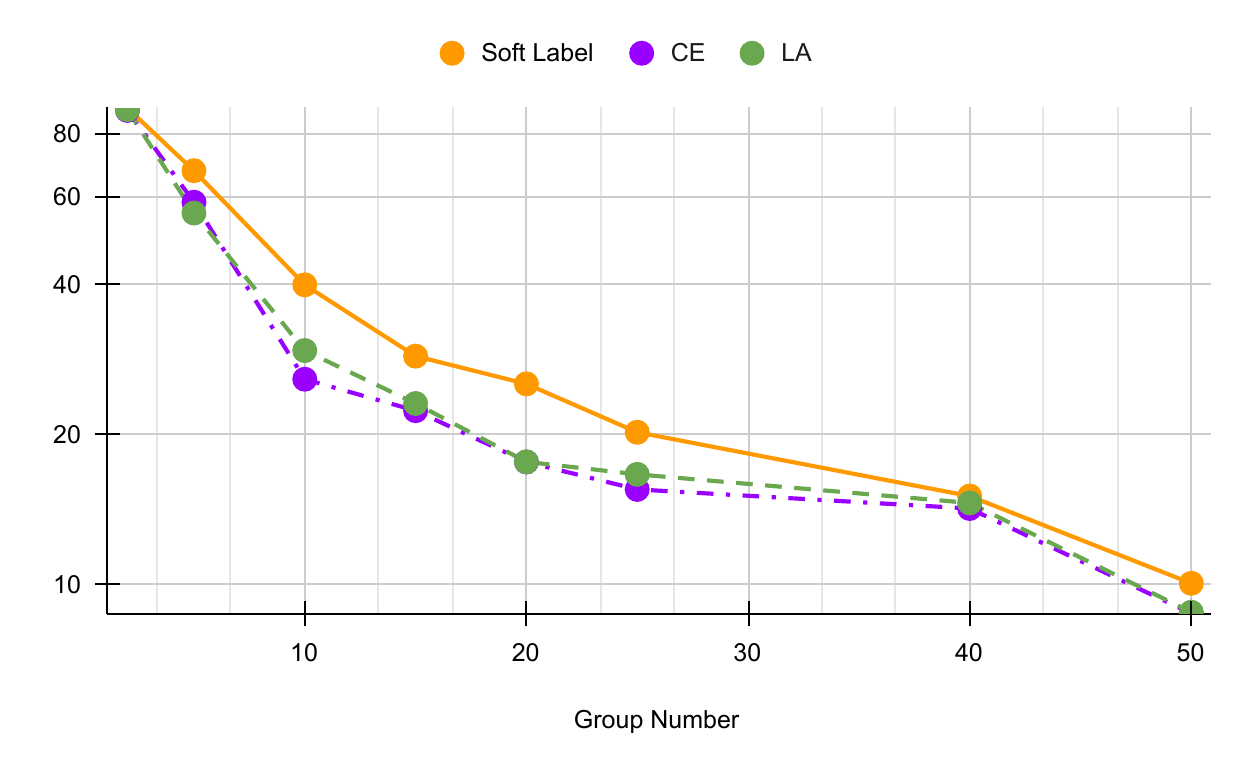}
    \caption{Comparison between our Soft labeling/CE/LA between the criterion of classification of AgeDB-DIR on the different number of groups, Y : Group prediction accuracy.} \label{group_cri}
\end{figure}

Another interesting observation from Fig.\ref{absolute} arises that, directly using the CE and LA would also make the predictions in the tail groups almost fail, that is because the tail groups are always the minorities.
Also, our Soft labeling can capture the information from other groups to help minorities in group imbalance.
As in Fig.\ref{group_cri}, the classification performance of the soft labeling is consistently better than that of the CE and LA, such as in 20 groups, the soft labeling has a 8\% improvement compared with others, which shows that the label similarity is also one very important characteristic in DIR and leverage the group similarities as that of the label similarities can help to take advantages of classification in helping DIR as \cite{8953836, pintea2023step}.

\begin{figure}[htbp]
    \centering
    \includegraphics[width=0.48\textwidth]{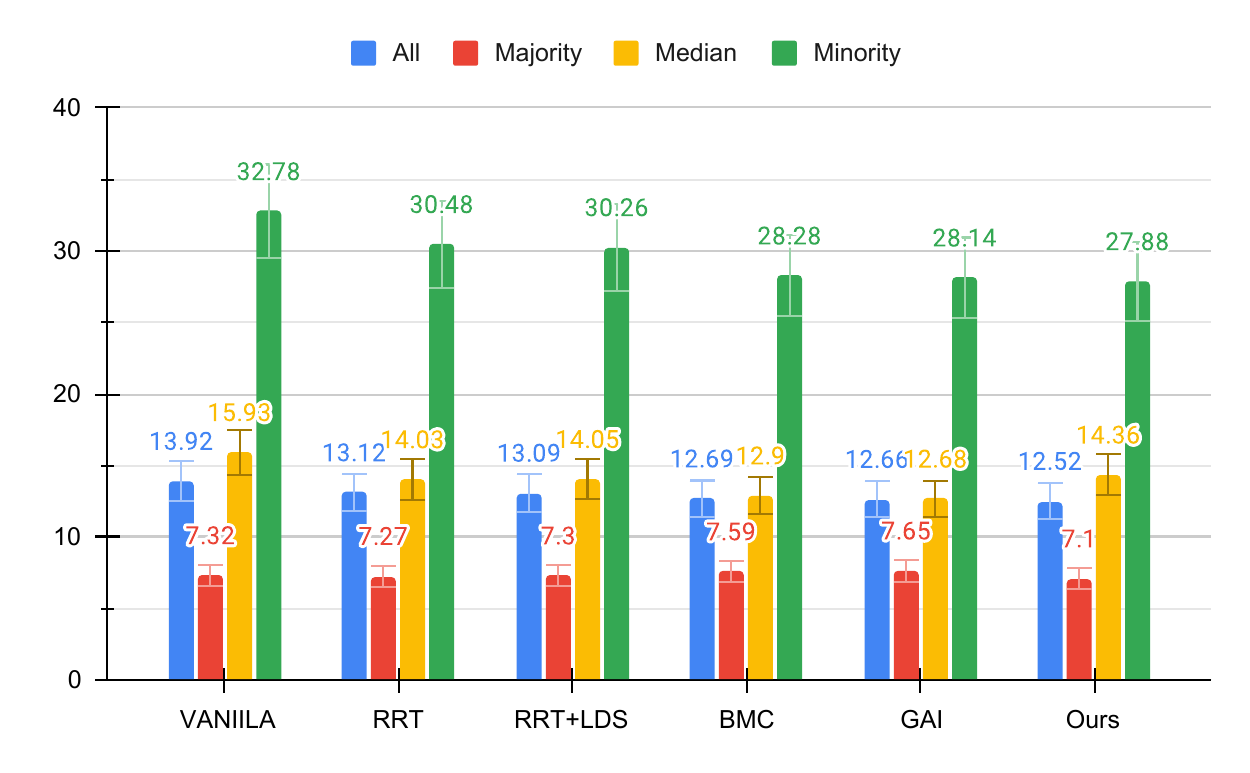}
    \caption{Comparison of MAE between various DIR solutions of b-MAE Results in Majority, Median, and Minority on IMDB-WIKI-DIR.} \label{bMAE}
\end{figure}

\textbf{IMDB-WIKI-DIR}: In the dataset IMDB-WIKI-DIR, which is also the largest DIR real-world dataset, our overall performance in Tab.\ref{imdb-table} achieved a satisfying result and is better than the current SOTA, specifically, we show that our method can have a better performance on the median and the few shot, it shows that our proposed method can exploit more information on the median and the few shots with the soft-label, resulting in an overall performance improvement.
Moreover, we show that our result in GM is also better than others both in the Med. and Few., showing a consistent superiority of our work over others and validating that our proposed method can better address the Med. and Few shots in the extreme case of imbalance.
As we can observe from Fig.\ref{bMAE}, the comparison of b-MAE results between the state-of-art DIR solutions and our method (20 groups) also validates the effectiveness of our method, showcasing our method can have a better performance on the balanced sets, especially on the minorities.

\textbf{STS-B-DIR}: In the dataset STS-B-DIR, it is easy to observe that the under-represented median data samples significantly outperform others in Tab.\ref{sts}, contributing to the overall performance enhancement.
More importantly,s our work simultaneously improved all shots performance in Pearson Correlation compared to others, that is because our soft labeling can help us to preserve the data similarity as that of labels in representation learning, which can enhance the correlations of the data and consistent with the smoothing-based methods (e.g. as VIR \cite{wang2023variational} and FDS and LDS \cite{yang21devoling}).

\begin{figure}[htbp]
    \centering
    \includegraphics[width=0.48\textwidth]{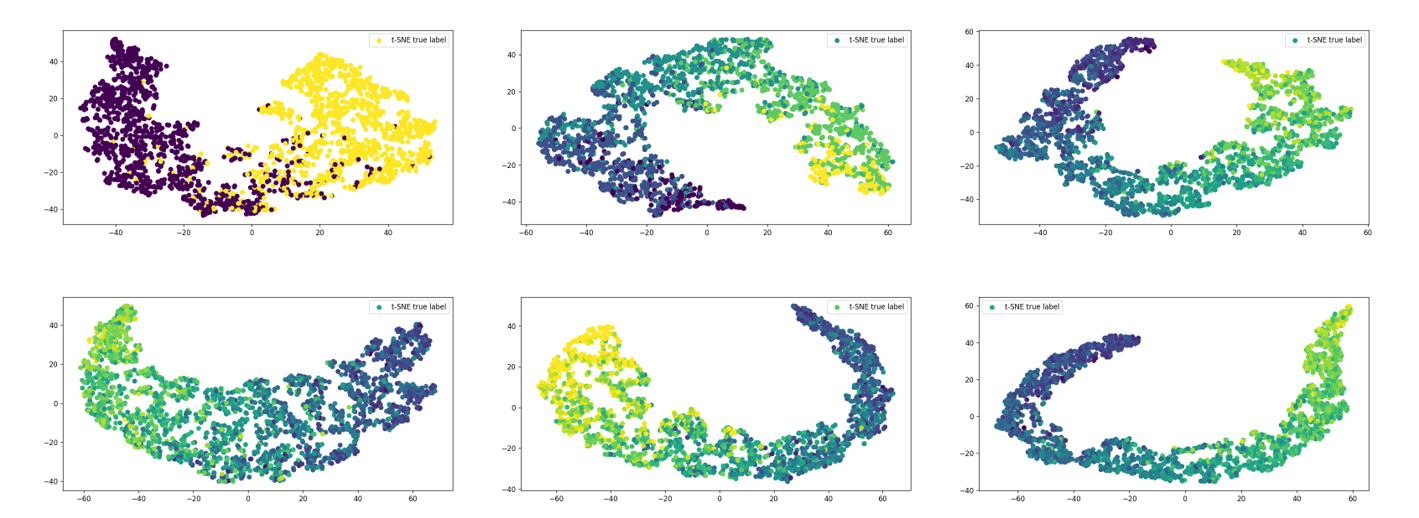}
    \caption{Comparison of TSN-E for the different number of groups on AgeDB-DIR, first row from left to right: 2/5/10 groups, second row from left to right: 20/40/50 groups.} \label{tsne}
\end{figure}

Furthermore, we show the T-SNE of the feature space under three different classification criteria, CE, LA, and Ours of AgeDB-DIR.
More importantly, as we can find from Fig.\ref{tsne}, with the increasing of the group numbers, we can observe that the feature space shows the same continuity as that of the label space (even in the case of 2 groups), demonstrating that our proposed method can preserve the ordinal of the label space.
As stated in \cite{gong2022ranksim, zha2023rank}, keeping the same structural between the feature and label space can be beneficial in addressing DIR, thus, the T-SNE in Fig.\ref{tsne} also explained the effectiveness of our proposed method.
Also, Fig.\ref{tsne} shows the continuity of the label space, more importantly, the distinctions of the features in different groups can be observed, which validates the effectiveness of our proposed group level contrastive learning.

\begin{table}[t]
\centering
\makeatletter\def\@captype{table}
\caption{Evaluation on AgeDB-DIR.}
\resizebox{0.5\textwidth}{!}{
\begin{tabular}{l|c|c|c|c|c|c|c|c}\hline
    \multirow{2}{*}{\diagbox{Method}{Shot}} & \multicolumn{4}{|c}{MAE$\downarrow$} & \multicolumn{4}{|c}{GM$\downarrow$} \\
    \cline{2-9}
    & All & Many. & Med. & Few. & All & Many. & Med. & Few. \\
    \hline
    VANILLA & 7.77 & 6.62 & 9.55 & 13.67 & 5.05 & 4.23 & 7.01 & 10.75 \\
    \hline
    SMOTER \cite{torgo2013smote} & 8.16 & 7.39 & 8.65 & 12.28 & 5.21 & 4.65 & 5.69 & 8.49 \\
    \hline
    SMOGN \cite{branco2017smogn} & 8.26 & 7.64 & 9.01 & 12.09 & 5.36 & 4.90 & 6.19 & 8.44 \\
    \hline
    RRT \cite{rrt} & 7.74 & 6.98 & 8.79 & 11.99 & 5.00 & 4.50 & 5.88 & 8.63 \\
    \hline
    RRT+LDS \cite{yang21devoling} & 7.72 & 7.00 & 8.75 & 11.62 & 4.98 & 4.54 & 5.71 & 8.27 \\
    \hline
    FOCAL-R \cite{lin2017focal} & 7.64 & 6.68 & 9.22 & 13.00 & 4.90 & 4.26 & 6.39 & 9.52 \\
    \hline
    SQINV \cite{yang21devoling} & 7.81 & 7.16 & 8.80 & 11.20 & 4.99 & 4.57 & 5.73 & 7.77 \\
    \hline
    SQINV + LDS \cite{yang21devoling} & 7.67 & 6.98 & 8.86 & 10.89 & 4.85 & 4.39 & 5.80 & 7.45 \\
    \hline
    LDS+FDS \cite{yang21devoling} & 7.55 & 7.01 & 8.24 & 10.79 & 4.72 & 4.36 & 5.45 & 6.79 \\
    \hline
    VAE \cite{vae} & 7.63 & 6.58 & 9.21 & 13.45 & 4.86 & 4.11 & 6.61 & 10.24 \\
    \hline
    DER \cite{amini2020deep} & 8.09 & 7.31 & 8.99 & 12.66 & 5.19 & 4.59 & 6.43 & 10.49 \\
    \hline
    Con-R \cite{keramati2023conr} & 7.20 & 6.50 & 8.04 & 9.73 & 4.59 & 3.94 & 4.83 & 6.39 \\
    \hline
    RankSim \cite{gong2022ranksim} & 7.02 & 6.49 & 7.84 & 9.68 & 4.53 & 4.13 & 5.37 & 6.89 \\
    \hline
    VIR \cite{wang2023variational} & 6.99 & \textbf{6.39} & 7.47 & \textbf{9.51} & 4.41 & 4.07 & 5.05 & \textbf{6.23} \\
    \hline
    LDS+FDS+DER \cite{wang2023variational} & 8.18 & 7.44 & 9.52 & 11.45 & 5.30 & 4.75 & 6.74 & 7.68 \\
    \hline
    Ours & \textbf{6.87} & 6.54 & \textbf{6.96 }& 9.83 & \textbf{4.30} & 4.10 & \textbf{4.39} & 6.45  \\
    \hline
\end{tabular}
}
\label{agedb_tabel}
\end{table}

\begin{table}[t]
\centering
\makeatletter\def\@captype{table}
\caption{Evaluation on IMDB-WIKI-DIR.}
\resizebox{0.5\textwidth}{!}{
\begin{tabular}{l|c|c|c|c|c|c|c|c}\hline
    \multirow{2}{*}{\diagbox{Method}{Shot}} & \multicolumn{4}{|c}{MAE$\downarrow$} & \multicolumn{4}{|c}{GM$\downarrow$} \\
    \cline{2-9}
     & All & Many. & Med. & Few. & All & Many. & Med. & Few. \\
    \hline
    VANILLA & 8.06 & 7.23 & 15.12 & 26.33 & 4.57 & 4.17 & 10.59 & 20.46 \\
    \hline
    SMOTER \cite{torgo2013smote} & 8.14 & 7.42 & 14.15 & 25.28 & 4.64 & 4.30 & 9.05 & 19.46 \\
    \hline
    SMOGN \cite{branco2017smogn} & 8.03 & 7.30 & 14.02 & 25.93 & 4.63 & 4.30 & 8.74 & 20.12 \\
    \hline
    SMOGN + LDS \cite{yang21devoling} & 8.02 & 7.39 & 13.71 & 23.22 & 4.63 & 4.39 & 8.71 & 15.80 \\
    \hline
    RRT+LDS \cite{yang21devoling}  & 7.79 & 7.08 & 13.76 & 24.64 & 4.34 & 4.02 & 8.72 & 16.92 \\
    \hline
    SQINV+LDS \cite{yang21devoling} & 7.83 & 7.31 & 12.43 & 22.51 & 4.42 & 4.19 & 7.00 & 13.94 \\
    \hline
    FOCAL-R+LDS \cite{yang21devoling} & 7.90 & 7.10 & 14.72 & 25.84 & 4.47 & 4.09 & 10.11 & 19.14 \\
    \hline
    BMC\cite{ren2022balanced} & 8.08 & 7.52 & 12.47 & 23.29 & - & - & - & - \\
    \hline
    GAI\cite{ren2022balanced} & 8.12 & 7.58 & 12.27 & 23.05 &  - & - & - & - \\
    \hline
    VAE  \cite{vae} & 8.04 & 7.20 & 15.05 & 26.30 & 4.57 & 4.22 & 10.56 & 20.72 \\
    \hline
    DER \cite{amini2020deep} & 7.85 & 7.18 & 13.35 & 24.12 & 4.47 & 4.18 & 8.18 & 15.18 \\
    \hline
    Con-R \cite{keramati2023conr} & 7.33 & 6.75 & 11.99 & 22.22 & 4.02 & 3.79 & 6.98 & 12.95 \\
    \hline
    RankSim \cite{gong2022ranksim} & 7.50 & 6.93 & 12.09 & 21.68 & 4.19 & 3.97 & 6.65 & 13.28 \\
    \hline
    VIR \cite{wang2023variational} & \textbf{7.19} & \textbf{6.56} & 11.81 & 20.96 & \textbf{3.85} & \textbf{3.63} & 6.51 & 12.23 \\
    \hline
    LDS + FDS + DER \cite{wang2023variational} & 7.24 & 6.64 & 11.87 & 23.44 & 3.93 & 3.69 & 6.64 & 16.00 \\
    \hline
    Ours & 7.22 & 6.71 & \textbf{11.42} & \textbf{20.25} & 3.88 & 3.68 & \textbf{5.74} & \textbf{11.13} \\
    \hline
\end{tabular}
}
\label{imdb-table}
\end{table}

\subsection{Ablation Study on Group Numbers}

\begin{figure}[htbp]
    \centering
    \includegraphics[width=0.48\textwidth]{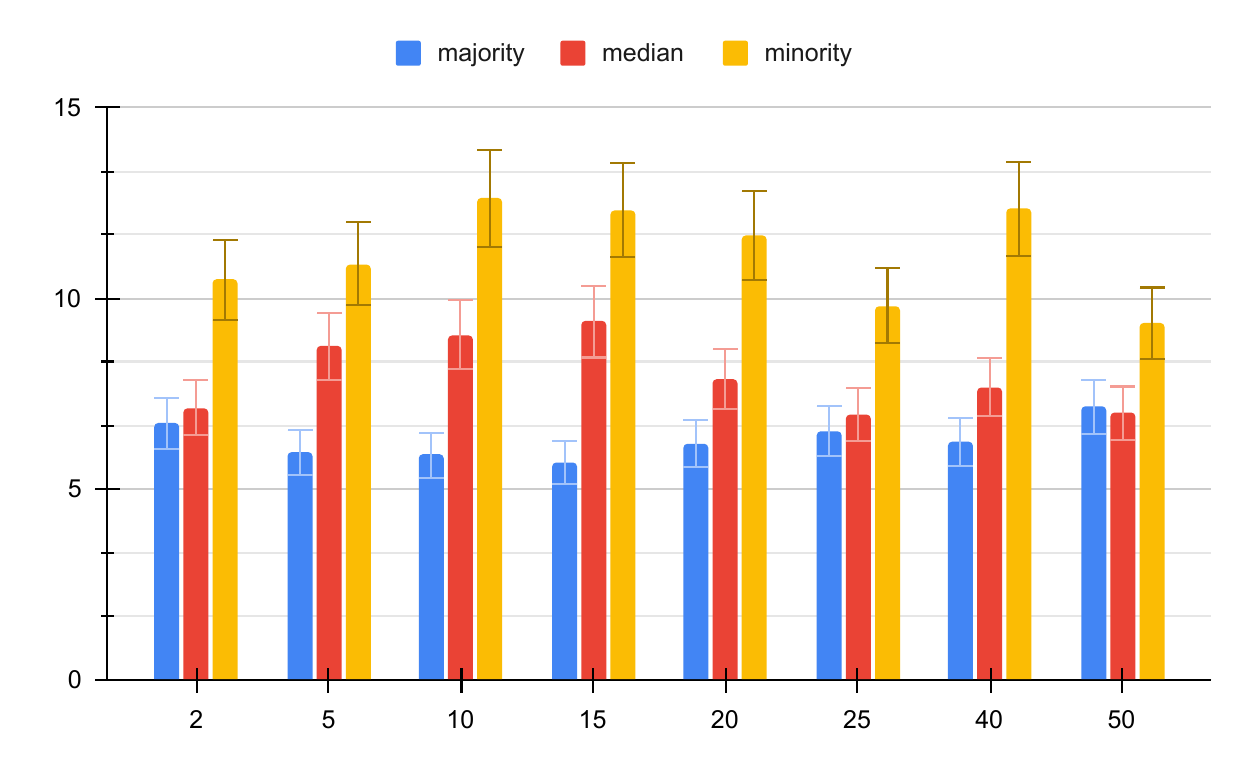}
    \caption{Comparison on Group Numbers \textbf{vs} MAE in Majority, Median and Minority. Y: MAE, X:group numbers.} \label{groups_mae}
\end{figure}

We also provide a detailed ablation study on the group numbers with the group prediction accuracy and the MAE on AgeDB in Fig.\ref{group_cri} and Fig.\ref{groups_mae}.
With the increasing of the group numbers, the prediction accuracy gradually drops, the reason why this phenomenon occurs comes from the data dependence over the groups. 
Therefore, we proposed soft labeling which can leverage the data dependence across the groups and yield a satisfying outcome.

Also, the increasing of the group numbers in Fig.\ref{tsne} also shows a more obvious continuity in the feature space, demonstrating that our method can have a better preservation of the feature continuity.
In Fig.\ref{groups_mae}, we can observe that each portion of data (Majority, Median, and Minority) varies slightly with the increasing of group numbers. 
Specifically, in 15, 20, 25, and 40 group settings, the performance of the majority shots is always close to each other while the median is varied slightly. 
Meanwhile, most of them always outperform other DIR solutions in Tab.\ref{agedb_tabel} and Fig.\ref{groups_mae}, which also shows the prominence of our proposed method.

\begin{table}
\caption{Evaluation on STS-B-DIR.}
\resizebox{0.5\textwidth}{!}{
\begin{tabular}{l|c|c|c|c|c|c|c|c}\hline   
    \multirow{2}{*}{\diagbox{Method}{Shot}} & \multicolumn{4}{|c}{MSE$\downarrow$} & \multicolumn{4}{|c}{Pearson Correlation$\uparrow$} \\
    \cline{2-9}
    & All & Many. & Med. & Few. & All & Many. & Med. & Few. \\
    \hline
    VANILLA & 0.974 & 0.851 & 1.520 & 0.984 & 74.2 & 72.0 & 62.7 & 75.2 \\
    \hline
    SMOTER \cite{torgo2013smote} & 1.046 & 0.924 & 1.542 & 1.154 & 72.6 & 69.3 & 65.3 & 70.6 \\
    \hline
    SMOGN \cite{branco2017smogn} & 0.990 & 0.896 & 1.327 & 1.175 & 73.2 & 70.4 & 65.5 & 69.2 \\
    \hline
    SMOGN + LDS \cite{yang21devoling} & 0.962 & 0.880 & 1.242 & 1.155 & 74.0 & 71.5 & 65.2 & 69.8 \\
    \hline
    RRT \cite{rrt} & 0.964 & 0.842 & 1.503 & 0.978 & 74.5 & 72.4 & 62.3 & 75.4 \\
    \hline
    RRT + LDS \cite{yang21devoling} & 0.916 & 0.817 & 1.344 & 0.945 & 75.7 & 73.5 & 64.1 & 76.6 \\
    \hline
    FOCAL-R \cite{lin2017focal} & 0.951 & 0.843 & 1.425 & 0.957 & 74.6 & 72.3 & 61.8 & 76.4 \\
    \hline
    INV \cite{yang21devoling} & 1.005 & 0.894 & 1.482 & 1.046 & 72.8 & 70.3 & 62.5 & 73.2 \\
    \hline
    INV + LDS \cite{yang21devoling} & 0.914 & 0.819 & 1.31 & 0.95 & 75.6 & 73.4 & 63.8 & 76.2 \\
    \hline
    VAE \cite{vae} & 0.968 & 0.833 & 1.511 & 1.102 & 75.1 & 72.4 & 62.1 & 74.0 \\
    \hline 
    LDS + FDS \cite{yang21devoling} & 0.907 & 0.802 & 1.363 & 0.942 & 76.0 & 74.0 & 65.2 & 76.6 \\
    \hline
    DER \cite{amini2020deep} &  1.001 & 0.912 & 1.368 & 1.055 & 73.2 & 71.1 & 64.6 & 74.0 \\
    \hline
    RankSim \cite{gong2022ranksim} & 0.903 & 0.908 & 0.911 & 0.804 & 75.8 & 70.6 & 69.0 & 82.7 \\
    \hline
    VIR \cite{wang2023variational} & 0.892 & \textbf{0.795} & 0.899 & 0.781 & \textbf{77.6}  & \textbf{75.2} & 69.6 & 84.5 \\
    \hline 
    LDS + FDS + DER \cite{wang2023variational} &  1.007 & 0.880 & 1.535 & 1.086 & 72.9 & 71.4 & 63.5 & 73.1 \\
    \hline
    Ours & \textbf{0.887} & 0.897 & \textbf{0.891} & \textbf{0.779} & 77.4 & 74.9 & \textbf{70.7} & \textbf{85.8}\\
    \hline
\end{tabular}
}
\label{sts}
\end{table}

\section{Conclusion}
In this work, we present a symmetric descending Soft labeling guided group-aware ordinal contrastive learning framework to learn a high-quality representation that both exhibits discriminative and similar characteristics simultaneously to address the DIR in a divide-and-conquer manner motivated by our theoretical analysis.
Extensive experiments on various real-world datasets verify the superiority of our method.
Our analysis of the results further validates the effectiveness of our proposed method.

\section{Appendix}

\section{Related work}
\label{sec:Related work}

\subsection{Imbalanced Classification}
Imbalance classification has been well explored by many researchers in the past few years.
Specifically, the training and testing set are drawn from different distributions, learning between different distributions are also well discussed in the previous literature \cite{pmlr-v206-chen23h, 10102307, 8953836}.
It can be divided into several categories. First of all, re-weighting \cite{reweight11, huang2016learning,reweight22, branco2017smogn} and re-sampling \cite{resam10,resam14, resam27, zhu2023nus}, is the most popular method in dealing with class imbalance. 
Meanwhile, post-hoc methods \cite{menon2021longtail,post43,post40} provide us a different perspective on imbalance class problems, they modified the output logits from the class prior to redeeming the imbalance.
Data augmentation methods \cite{data157, data160,data49,shi2022improving} such as mixup, have been introduced to address the imbalance in classification as they can increase the quantity of data and improve the quality of data under class imbalance.

Furthermore, learning a robust representation is also an effective way to handle the class imbalance, as in \cite{huang2016learning,Li_2022_Supervised}, supervised contrastive Learning has been introduced to the imbalance problems, and also achieved satisfying performance.
\cite{Yu2022Re-Balancing} tried to leverage the instance difficulty to measure the imbalance and integrated it into the re-balance strategy.
Moreover, emerging techniques have been proposed to address the imbalance, such as Mixture-of-Experts \cite{zhou2022mixture}, feature selection \cite{han2022locating} etc.
Besides, as \cite{liu2021self} pointed out, high-quality representation can also be robust to class imbalance. 
Thus, researchers tried to learn a robust representation by improving the feature quality \cite{rep123, CCGC}, leveraging metric learning \cite{rep50}, and using prototype learning \cite{rep15}.
As a result, the high-quality feature representation helps the model to perform robustly under class imbalance.
Therefore, in this work, we explore how to leverage the benefits from the classification to help DIR.

\subsection{Deep Imbalanced Regression}
Deep imbalance regression \cite{zhang2023deep} an under-explored porblem in machine learning. \cite{torgo2013smote} tried to use synthetic data for the imbalance. \cite{REBAGG} used an ensemble method for the regressors. \cite{Density-based-weighting} used a re-weight-based method based on the estimated distribution. 
\cite{jiang2023mixture, wu2023mixup} also adopted an uncertainty-based mixture of experts for dealing with imbalance regression.
\cite{yao2022c} introduced Mixup to address the regression problems.
\cite{xiong2023deep} used hierarchical classification adjustment with distillation for the imbalance regression problems.
\cite{wang2023variational} also leveraged the neighboring similar samples to conduct variational inference in DIR.
\cite{yang21devoling} adopt label/feature smoothing to deal with the imbalance.
However, smoothing \cite{yang21devoling} which is used on either feature or label space to borrow the neighboring knowledge and performed at each sample
Differently, our work is to aggregate the knowledge from a group level, which can better leverage the advantages from classification as that of \cite{pintea2023step, CCGC} and also take advantage of the knowledge from all groups.

Also, several methods have been proposed to leverage the robust representation to address the imbalance regression.
\cite{Mean-Variance, pintea2023step} tried to use the classification loss to solve the regression tasks. 
\cite{rank-consistent} used ordinary classification method to deal with regression. 
\cite{zha2022supervised, keramati2023conr, wang2023fend}, but also maintains the feature similarity to redeem the imbalance in regression.
\cite{zhang2023improving} explained that directly minimizing mean square error loss does not increase the marginal entropy, thus they introduced the ordinary entropy to DIR problems to a more robust feature.
Furthermore, \cite{silva2022model} studied the effectiveness of metric Squared Error Relevance Area (SERA) in imbalance regression.
\cite{stocksieker2023data} combined a weighted resampling and a data augmentation method in a regression framework.
\cite{LiangBilateral} employed the cumulative learning from the feature learning to the regressor learning.
\cite{xiong2024} proposed a hierarchical classification adjustment for DIR but ignored the intrinsic characteristic of label similarity in DIR.
Among them, \cite{gong2022ranksim} is a relevant paper compared with us. 
However, rank similarity regularization is used to exploit the discriminative information to make each feature representation distinguishable rather than considering the similarity information across data. 
Our works addressed the DIR by explicitly leveraging the advantages of the classification, which is different from the current implicit realization solutions.

\subsection{Model architecture}
\begin{figure}
  \centering
  \includegraphics[width=1\linewidth]{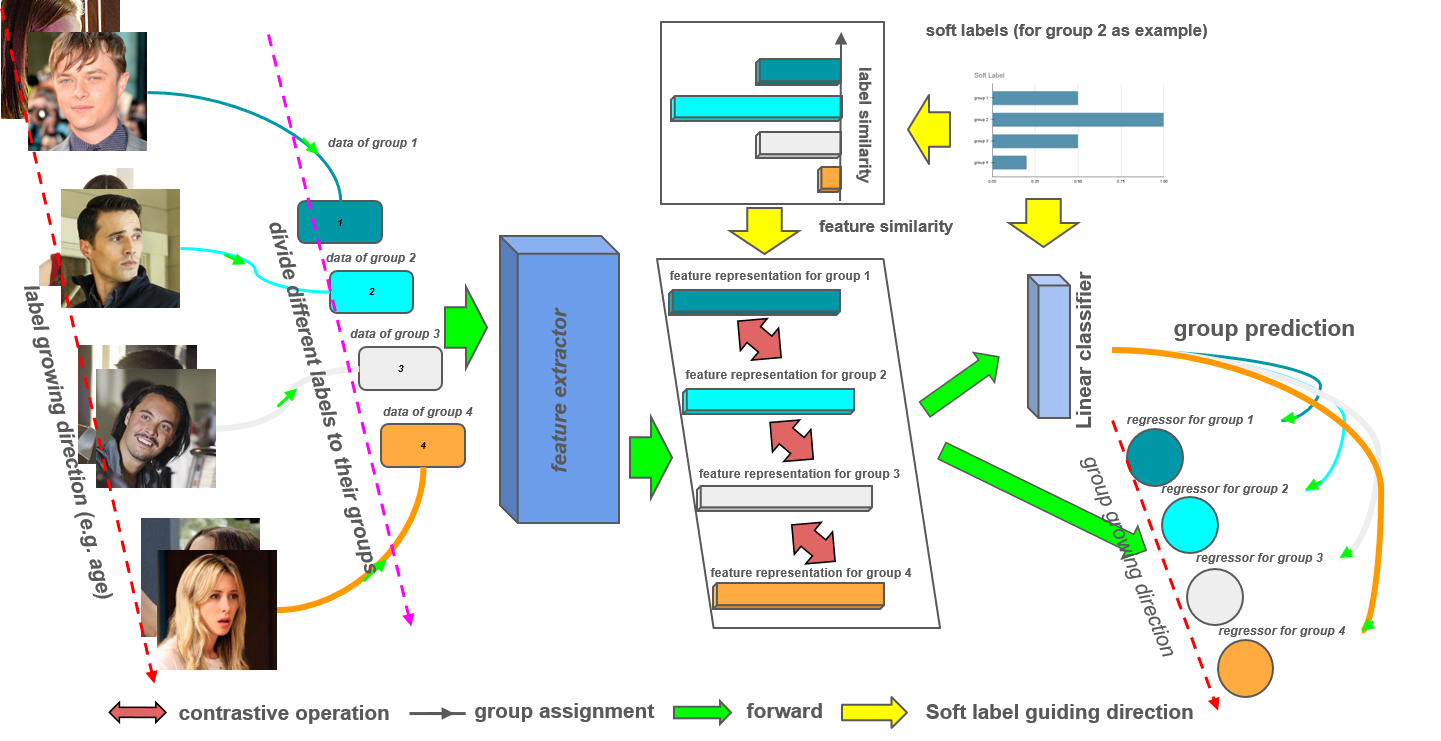}
  \caption{Demonstration of our method in a 4-group-division for the imbalanced age-estimation dataset of DIR, the same color of the line and the circle denote the same group, we show a four-group division of our proposed method in this architecture.} \label{arch}
\end{figure}
The architecture of our proposed method is in Fig. \ref{arch}.
We show that we address the DIR in a divide-and-conquer manner.
Firstly, data samples are grouped into different groups according to their labels.
Then, all of them are sent to a feature encoder and obtain their features.

\textbf{(Divide)} All of the samples go to a classifier and get a corresponding group prediction for each data sample.

\textbf{(Conquer)} In training, the group prediction would be used for a classification loss.
And each MSE for each data sample is calculated on each expert regressor given their ground truth group index.
In testing, the group prediction is used to identify which expert regressor the data sample should be sent to in oroder to conduct the regression.

\subsection{Additional Implementation details}
To make a fair comparison, the vanilla model we used in our experiments is ResNet-50, also, we take ResNet-50 as our backbone and we revised the output dimension from 1-d to a classification head and a multi-expert regressor for our method. 
Also, we use Mean square error (MSE) and Mean absolute error (MAE) as the loss function. 
Adam is employed as the optimizer, the momentum is set as 0.9 and the weight decay is 1e-4.
Temperature parameter is set to 2.5 by conducting a grid search from 1 to 5, as we recommend $\geq$ 1 temperature parameter for the increasing of batch size (our batch size is 128 due to the GPU limitations).
We pre-process the data during training to fully exploit the potential of the data, which is also a standard setting in the previous works.
We trained the classification \& regression head under the grid search between 50 to 90 epochs, we suggest different epochs for the different number of groups.
Therefore, we do not need balance the three losses in the final loss in the fine tuning.
Following \cite{yang21devoling}, we report MAE and Geometric mean (G-Mean) for AgeDB-DIR and IMDB-WIKI-DIR, MSE and Pearson Correlation for the STS-B-DIR.
Following \cite{ren2022balanced}, we reported the balanced-MAE (b-MAE) for IMDB-WIKI-DIR.

To show the effectiveness of our method compared with other baseline methods, following \cite{yang21devoling}, we present several classical imbalance solutions :
1) Synthetic data method: we present \cite{branco2017smogn} and \cite{torgo2013smote} which created synthetic data from minority to majority to resolve the imbalance.
2) Focal Loss R : based on \cite{lin2017focal}, \cite{yang21devoling} proposed Focal-R which modified the scaling factor.
3) Decoupled training, a regressor re-training (RRT), which re-trains the regressor while freezing the well-trained encoder.
4) Balanced MSE, which revised MSE loss from a statistical view, namely GMM-based Analytical Integration (GAI) and Batch-based Monte-Carlo (BMC), to measure the imbalance in the training and inference stage, we also compare our work with other state-of-art methods in DIR.
Every experiment has been replicated with five random seeds and the reported result is the average of them, also, same dataset shares the same experiment setting in this paper.

\subsection{Demonstration of Symmetric descending Soft label}
\begin{figure}
    \centering
    \includegraphics[width=\linewidth]{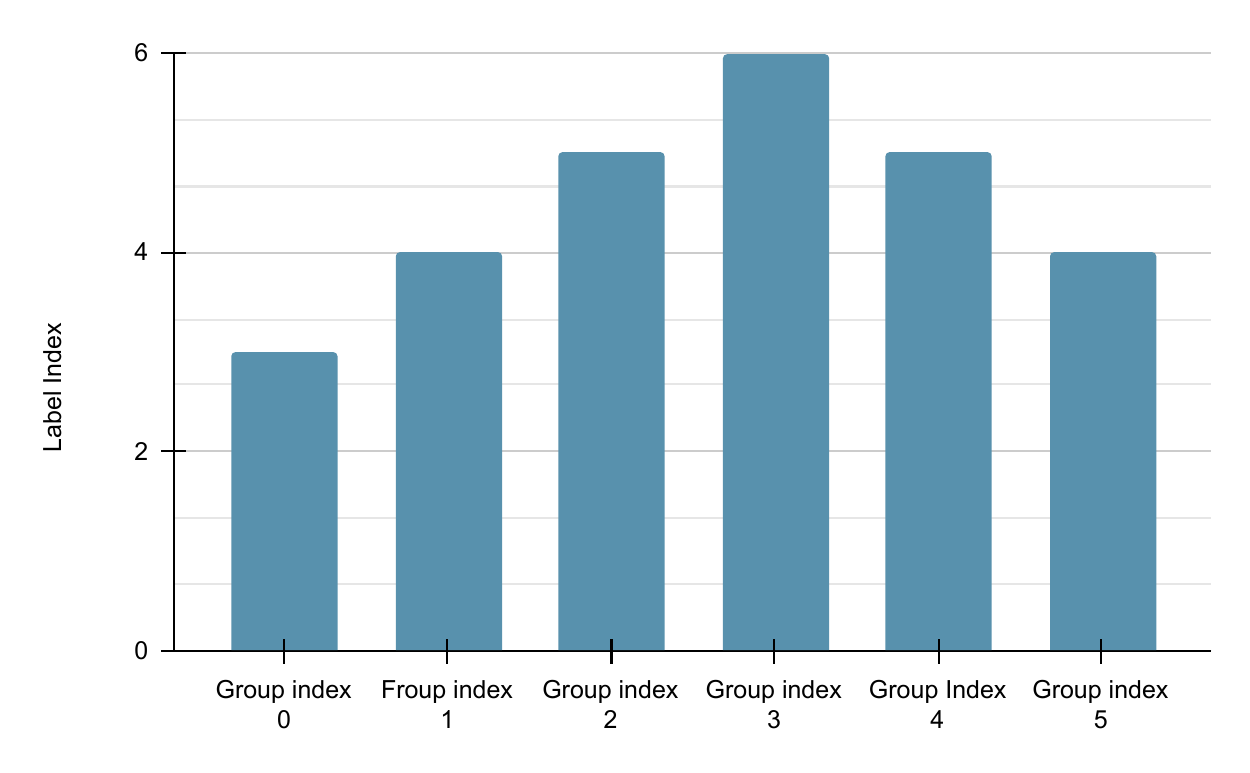}
    \caption{A demo for the Symmetric Descending Soft Labels}
    \label{softs}
\end{figure}

We show a demonstration for our proposed Symmetric descending Soft label in Fig.\ref{softs}.
In Fig.\ref{softs}, assuming the ground truth class index is group 3 (for a 6 group classification and $\beta=1$), the soft label is encoded in a symmetric and descending manner.

\renewcommand{\floatpagefraction}{.9}
\begin{figure}[htbp]
    \centering
    \includegraphics[width=0.48\textwidth]{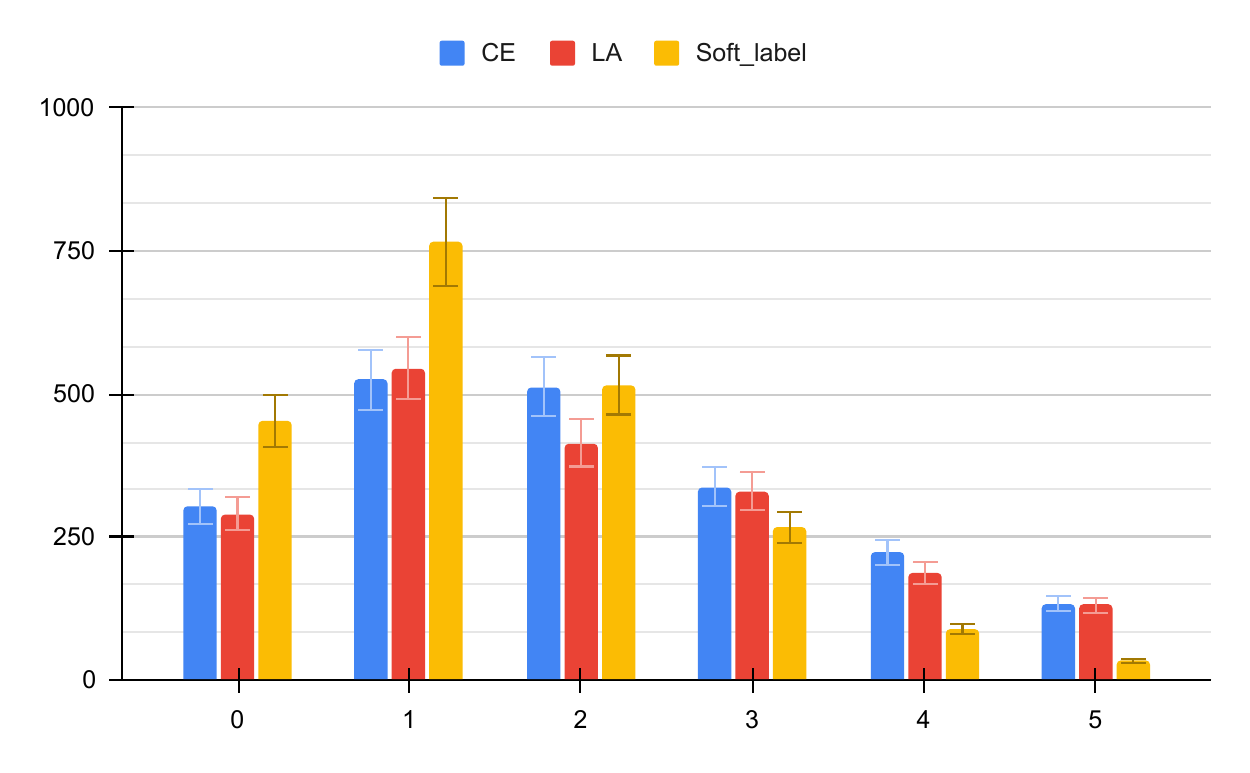}
    \caption{Comparison of absolute distance between Soft CE/LA/Soft label of AgeDB-DIR. From left to right, LA, CE, and Ours. X-axis : number of groups, Y-axis : number of samples.} \label{tsne_abs}
\end{figure}

\subsection{Ablation Study and analysis on Classification Criterion}

We provide the ablation study over the criterion of the classification, as we test three different criteria, soft label, cross-entropy, and logit adjustment \cite{menon2021longtail} in Table.\ref{ab_age_mae}.
As we can find from Table, we show that same as the ablation study of different group numbers and group estimation accuracy, the accuracy of group prediction gradually drops with the increasing of group numbers under the case of data dependence. 
However, the group prediction trained under the soft label always outperforms the other two criteria, which proves that preserving the group-wise similarities in the representation learning is crucial, as we can leverage the similar features from the majorities to redeem the minorities.
The ablation study on our proposed loss in Table \ref{criterion} also validates the effectiveness of our method.

As we can observe from Table \ref{criterion}, when we compare our proposed method even with the state-of-art imbalance classification solutions, our soft label can also outperform than others, that is because in DIR, the label space preserve the label ordinal and label similarity.
Different from the classification where the label space are distinct and non ordinal, soft label in the DIR can better capture these characteristics, which would result in a performance improvement in the group classification.
In Fig.\ref{tsne_abs}, we show that the soft label can help the model to reduce the distance from mis-classifying the samples to the close-groups, as it can help the model to maintain the similarities across the groups.
Results of the Soft label also shows the effectiveness of label smoothing in addressing imbalance as stated in \cite{shwartz-ziv2023simplifying}.

\begin{figure}[htbp]
    \centering
    \includegraphics[width=0.48\textwidth]{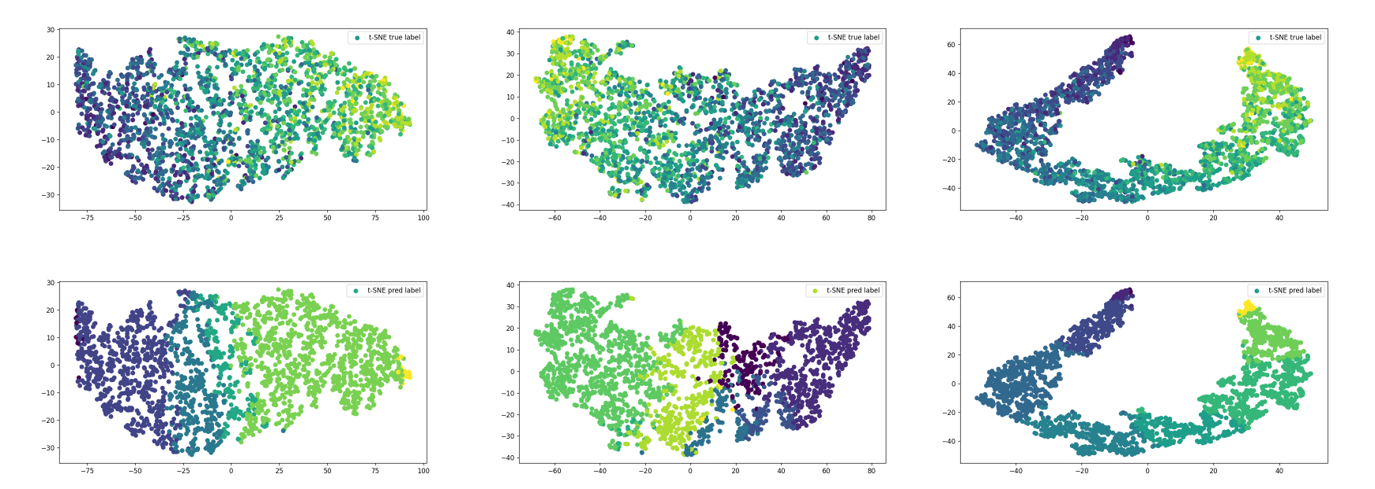}
    \caption{Comparison of TSN-E between Soft CE/LA/Soft label of AgeDB-DIR. From left to right, LA, CE, and Ours. First row: ground truth group. Second row: predicted group.} \label{tsne_group}
\end{figure}

Lastly, if we only use the soft label, we show that soft label can also help the DIR, which is in alignment with the findings in \cite{pintea2023step}, and the soft label can better preserve the group similarities as that of in the label space.
Also, in Fig.\ref{tsne_group}, 
all three methods show an ordinal characteristic either from left to right or exhibit continuous.
The soft label can better preserve the feature space as the labels since it shows a strong continuity, which also validates the effectiveness of our proposed method.

\section{Future Work}
In future work, we need to better exploit the intrinsic nature of the data to improve the classification performance under data dependence across the groups.
We need to investigate how to balance the discriminative and similar characteristics to address the DIR problem.
Moreover, we will explore more on the connections between the classification and the regression, studying a qualitative connection between the imbalance classification and regression.

\begin{table} \label{groups}
\centering
\caption{Ablation study on \textbf{group numbers} of AgeDB-DIR.} 
\begin{tabular}{|c|c|c|c|c|}
    \hline
    Group num  & MAE  & G-mean & MAE-GT & G-Mean-GT \\ 
    \hline
    2  & 7.11 & 4.65 & 6.26 & 4.03 \\ 
    \hline 
    5  & 6.93 & 4.41 & 4.53 & 2.93 \\ 
    \hline
    10 & 7.07  & 4.57 & 3.38 & 2.14\\ 
    \hline
    15 & 6.98  & 4.48 & 2.15 & 1.20 \\ 
    \hline
    20 & 6.97  & 4.45 &  2.09 & 1.23 \\ 
    \hline
    25 & 6.87  & 4.30 & 1.45 & 1.11 \\ 
    \hline 
    40 & 7.01 & 4.55 & 1.84 & 0.87\\ 
    \hline
    50 & 7.29 & 4.63 & 2.47 & 0.97 \\ 
\hline
\end{tabular}
\label{ab_age_mae}
\end{table}

\begin{table}
  \caption{Comparison between different SOTA imbalance group (\#) classification (\%) solutions for AgeDB.}
  \label{cls}
  \resizebox{0.5\textwidth}{!}{
  \begin{tabular}{|c|c|c|c|c|}
    \hline
    \#&Soft Labels(Ours) & NC \cite{yang2022inducing} & DO \cite{Cong_Xuan_Liu_Zhang_Pagnucco_Song_2024} & ReBAT \cite{wang2024balance}\\
    \hline
    5  &  67.71 & 66.01 & 66.72 & 67.04 \\
    \hline
    10 & 39.99 & 38.75 & 38.41 & 39.29 \\
    \hline
    25 & 20.56 & 18.43 & 19.07 & 19.14 \\
    \hline
    50 & 10.09 & 9.67 & 9.44 & 9.78 \\
  \hline
\end{tabular}
}\label{criterion}
\end{table}

\begin{table}
  \caption{Comparison between hyper-parameter $\lambda_1$ and $\lambda_2$ on AgeDB-DIR (group 25)}
  \label{cls}
  \resizebox{0.5\textwidth}{!}{
  \begin{tabular}{|c|c|c|c|c|c|}
    \hline
    $\lambda_1$ & 0.1  &  0.5 & 1 & 2 & 5\\
    \hline
    MAE  &  7.17 & 6.87 & 6.94 & 7.26 & 7.44 \\
    \hline
    $\lambda_2$ & 0.1  &  0.5 & 1 & 2 & 5\\
    \hline
    MAE  &  6.97 & 6.92 & 6.87 & 7.01 & 7.24 \\
    \hline
\end{tabular}
}
\end{table}

\section{Acknowledgments}
We appreciate constructive feedback from anonymous reviewers and meta-reviewers. 
Thanks to Dr.Qi Chen for her valuable suggestions.
This work is supported by the Natural Sciences and Engineering Research Council of Canada (NSERC), Discovery Grants program.

\bibliography{aaai25}

\end{document}